# MedHopQA: A Disease-Centered Multi-Hop Reasoning Benchmark and Evaluation Framework for LLM-Based Biomedical Question Answering

Rezarta Islamaj[1], Robert Leaman[1], Joey Chan[2], Nicholas Wan[3], Qiao Jin[1], Natalie Xie[1], John Wilbur[1], Shubo Tian[1], Lana Yeganova[1], Po-Ting Lai[1], Chih-Hsuan Wei[1], Yifan Yang[1], Yao Ge[1], Qingqing Zhu[1], Zhizheng Wang[1], and Zhiyong Lu[1*]

[1]*National Library of Medicine, Division of Intramural Research, Bethesda, MD, US*
[2]*University of Illinois at Urbana-Champaign, Department of Computer Science, Urbana, IL, US*
[3]*University of Michigan Medical School, Ann Arbor, Michigan, US*

**Corresponding author: E-mail: Zhiyong.Lu@nih.gov*

*Keywords: multi-hop question answering, biomedical NLP, benchmark evaluation, LLM-as-a-judge, training data contamination*

*MedHopQA is a disease-centered multi-hop biomedical question answering benchmark and reusable construction framework. The dataset comprises 1,000 expert-curated, open-ended question–answer pairs built from Wikipedia article pairs and validated through a human–AI pipeline. Zero-shot evaluation of four frontier LLMs confirms multi-hop biomedical reasoning remains challenging.*

## Abstract

Evaluating large language models (LLMs) in the biomedical domain requires benchmarks that can distinguish genuine reasoning from pattern matching and that remain discriminative as model capabilities improve. Existing biomedical question answering (QA) benchmarks are limited in this respect: multiple-choice formats allow models to succeed by answer elimination rather than inference, and widely circulated exam-style datasets are subject to performance saturation and training data contamination. Multi-hop reasoning. i.e., the ability to integrate information across multiple sources to derive an answer, is central to clinically meaningful tasks such as diagnosis support, literature-based discovery, and hypothesis generation, yet remains underrepresented in current biomedical QA benchmarks.

We present MedHopQA, a disease-centered multi-hop reasoning benchmark of 1,000 expert-curated question–answer pairs introduced as a shared task at BioCreative IX. Each question requires synthesis of information across two distinct Wikipedia articles, and answers are provided in open-ended free-text format rather than as multiple-choice selections. Gold annotations are augmented with ontology-grounded synonym sets (MONDO, NCBI Gene, NCBI Taxonomy) to support both lexical and concept-level evaluation. The dataset was constructed through a multi-stage human–AI pipeline combining structured human annotation, triage, iterative verification, and LLM-as-a-judge validation. To reduce leaderboard gaming and contamination risk, the 1,000 scored questions are embedded within a publicly downloadable set of 10,000 questions, with answers withheld, on a CodaBench leaderboard.

Evaluation of four frontier LLMs under a zero-shot setting (GPT-5.1, Gemini 2.5 Pro, Claude Sonnet 4.5, and GPT-4o) reveals performance variation across answer types, with overall accuracy ranging from 66.3% to 83.4%. Performance is strongest on chemical and anatomical questions and most variable on disease and gene/protein categories, where fine-grained semantic discrimination is required.

MedHopQA provides both a benchmark and a reusable framework for constructing future biomedical QA datasets that prioritize compositional reasoning, saturation resistance, and contamination mitigation as design constraints.

## Introduction

Biomedical text mining has traditionally been organized around task-specific information extraction. Named entity recognition, relation extraction, event detection, and document classification have been developed and evaluated as discrete, pipelined subtasks, each with dedicated models, training data, and evaluation benchmarks [1-7]. This task-specific paradigm has supported large-scale biocuration efforts by enabling the systematic extraction of structured knowledge from the literature through pre-defined schemas [8]. In contrast, natural language question answering provides end users with a more flexible interaction, in which information needs are posed directly as ad-hoc questions and system responses returned in fluent natural language. The rise of large language models (LLMs) has made this setting increasingly practical, enabling systems to retrieve, combine, and interpret information to produce coherent answers [9-11]. However, natural language question answering exposes different aspects of system behavior, particularly the ability to reliably integrate information across sources and perform multi-step reasoning. As a result, this shift introduces new demands on dataset construction and evaluation, requiring benchmarks that capture not only factual correctness but – crucially – can reliably evoke and evaluate reasoning over biomedical knowledge.

The benchmarks that shaped a decade of biomedical NLP progress were designed for a different class of systems, and their structural limitations have become consequential as LLMs have matured. Four limitations are particularly relevant:

**Multiple-choice questions Format.** Most widely used benchmarks, such as MedQA (USMLE) [12], MedMCQA [13], PubMedQA [14], the medical subsets of MMLU [15], rely on multiple-choice questions (MCQ). MCQ scoring is deterministic and supports straightforward cross-model comparison, but makes it difficult to distinguish a model that pattern-matches to the correct answer from one that derives it through reasoning.

**Performance Saturation.** State-of-the-art LLMs now score above 95% on MedQA [16], approach expert-level performance on MMLU medical subsets[17, 18], and have largely exhausted the discriminative capacity of PubMedQA's three-class label space[19]. When frontier models cluster near the ceiling of a benchmark, it ceases to be informative about capability or failure modes.

**Training data contamination.** USMLE-style questions, Indian medical entrance exam banks, PubMed abstracts, and their derivatives are represented in the pretraining corpora of all major LLMs. There is mounting evidence that high scores on these benchmarks reflect, at least in part, near-neighbor retrieval from training data rather than compositional reasoning [20, 21].

**Limited reasoning depth.** Most previous biomedical QA datasets are designed for single-hop retrieval, where the answer can be located within a single document or passage. However, meaningful biomedical questions often require integrating information across sources. For example, determining whether a drug is contraindicated in a patient with a specific comorbidity may require integrating pharmacological mechanisms, disease pathophysiology, and interaction data. Similarly, questions about rare diseases may require chaining across gene function, phenotypes, and treatment evidence distributed across multiple sources. Such compositional reasoning is central to tasks such as clinical decision support, hypothesis generation, and literature-based discovery. Datasets for multi-hop reasoning (i.e., requiring inference across multiple documents or knowledge units) are therefore important for benchmarking the capabilities of deployed systems.

Motivated by these limitations, we present MedHopQA [22] to directly address the multi-hop gap with a dataset and evaluation framework designed for the current LLM evaluation landscape. To our knowledge, it is the first biomedical QA benchmark to combine explicit multi-hop structure, open-ended answer format, and community-scale evaluation. This combination responds to the format rigidity, saturation, contamination vulnerability, and shallow reasoning limitations identified in the related works.

This paper makes three primary contributions. First, we present the MedHopQA dataset: 1,000 expert-curated, multi-hop question–answer pairs introduced as a shared task at BioCreative IX. Questions are constructed from Wikipedia, which provides a semi-structured, multi-document knowledge sources that supports construction of multi-hop reasoning chains while remaining broadly accessible. Each question requires at least two reasoning hops across distinct Wikipedia articles, such that no single source contains a sufficient answer. Gold annotations include final answers and synonymous alternatives drawn from domain ontologies (MONDO, NCBI Gene, NCBI Taxonomy), supporting both lexical and concept-level evaluation.

Second, we describe a generalizable framework for QA dataset creation that integrates structured human annotation, AI-assisted augmentation, and multi-stage validation. While instantiated here for disease-centered biomedical questions sourced from Wikipedia, the pipeline (seed curation, AI augmentation, blind triage, iterative verification, and LLM-as-a-judge answer validation) is applicable to other biomedical subdomains, knowledge sources, and reasoning structures.

Third, the dataset is publicly available on CodaBench, where researchers can download questions, submit system outputs, and optionally participate in a public leaderboard. To prevent training data contamination, the 1,000 test questions are hidden within a larger set of 10,000 questions, and answers are not publicly available. MedHopQA, thus enables standardized community-wide comparison across systems.

These contributions sit at the intersection of two traditions: the biomedical text mining evaluation culture of BioCreative, which has driven rigorous community-wide assessment for two decades, and the emerging LLM evaluation literature, which has recognized the need for benchmarks able to discriminate between reasoning ability beyond the level that saturated MCQ tasks can provide.

## Related Work

Table 1 summarizes representative biomedical and clinical QA datasets along dimensions relevant to current LLM evaluation, including question format, saturation risk, and contamination risk. Saturation refers to the remaining discriminative headroom a benchmark provides for newly-developed models; contamination refers to the likelihood that benchmark instances, close paraphrases, or answer patterns have entered pretraining, instruction-tuning, or benchmark-optimization corpora. These factors are related but distinct, and benchmark utility for LLM evaluation depends jointly on task structure, source material, and exposure.

The broader landscape spans extraction-oriented benchmarks grounded in PubMed literature (BioASQ[23], PubMedQA [14], BioRead [24]), clinical text benchmarks derived from EHRs and structured records (emrQA [25], MIMIC-III QA / MedNLI [26], RadQA [27]), and a range of exam-based MCQ benchmarks (MedQA [12], MedMCQA [13], MMLU Medical [15], AfriMed-QA [28]). More recent datasets have shifted toward open-ended formats and richer evaluation signals: HealthBench [29] uses physician-authored scenarios with rubric-based scoring; K-QA [30] evaluates comprehensiveness and hallucination on patient-authored questions; MedREQAL [31] targets evidence-grounded synthesis from systematic reviews; and LongHealth [32] and ClinIQLink [33] stress-test long-context aggregation and multi-hop formulations respectively. The structural limitations of the MCQ-dominated subset of this landscape (e.g., saturation, contamination, and format-driven ambiguity about whether a correct answer reflects reasoning or pattern matching) are discussed in the Introduction.

The subset of benchmarks most relevant to MedHopQA are those that explicitly target multi-hop or multi-document reasoning. In the general domain, HotpotQA[34], MuSiQue[35], and 2WikiMultiHopQA[36] established the methodological foundations for multi-hop evaluation, including supporting-fact annotation and step-level verification. However, their applicability for biomedical LLM evaluation is limited by domain mismatch, and several have become targets for benchmark-specific optimization. Within the biomedical domain, MedHop[37] was the first benchmark designed explicitly for multi-document reasoning, using MedlinePlus and UMLS as source material. Its multiple-choice format, however, constrains evaluation scope and introduces answer cueing effects that can allow models to succeed without executing the intended reasoning chain. BioHopR [38] advances this line of work by isolating one-hop and two-hop reasoning over a biomedical knowledge graph (PrimeKG), with the consistent performance decline from one-hop to two-hop items confirming that structured multi-step reasoning remains a genuinely unsaturated evaluation dimension. ClinIQLink[33] includes explicit multi-hop and inverse multi-hop formulations grounded in verified clinical sources, though as a shared-task benchmark it will benefit from future hidden test-set refreshes to guard against leaderboard-targeted optimization.

MedHopQA addresses the gap that remains after these contributions. Relative to MedHop, it replaces the MCQ format with open-ended free-text generation, eliminating answer cueing and requiring models to produce rather than select the correct inferential output. Relative to BioHopR, it moves from knowledge-graph-constrained answer sets to free-text generation over multi-document Wikipedia article pairs, broadening the surface of reasoning that must be engaged. The human–AI–human construction pipeline and multi-stage LLM-as-a-judge validation framework are designed to ensure that: 1) each question is answerable only by synthesizing information across two distinct source documents, and 2) gold annotations are extended to include ontology-grounded synonym sets (MONDO, NCBI Gene, NCBI Taxonomy) to support robust concept-level evaluation. Because evaluation answers are withheld and the scored test items are embedded within a larger 10,000-question package, MedHopQA also provides structural resistance to the exact-instance contamination that affects long-circulating public exam datasets.

*Table 1 Major Biomedical and Clinical QA Datasets*

| Dataset | Domain | Source material | QA format | Size | MH / Sat / Cont | Notes |
|---|---|---|---|---|---|---|
| BioASQ [23] | Biomedical | PubMed articles, snippets | Y/N, factoid, list, summary; structured + free text | ~300-500/yr | ▲ ● ◆ | Format-rich; annual refresh |
| PubMedQA [14] | Biomedical | PubMed abstracts | Y/N/maybe; single label | ~1k | ▲ ● ◆ | Coarse label space, low headroom |
| BioRead [24] | Biomedical | Biomedical literature | Fill-in-the-blank; word/phrase | Variable | ▲ ● ◆ | Narrow task, weak diagnostic value |
| MEDIQA [39, 40] | Clinical/medical | Medical text, consumer questions | Entailment/QA/summarization; classification | ~5k+ | ▲ ● ◆ | Useful legacy inference benchmark |
| emrQA [25] | Clinical | EHR notes | Span QA; text spans | ~1M+ | ▲ ● ◆ | Authentic notes, but templated questions |
| MIMIC-III QA / MedNLI [26] | Clinical | ICU records | NLI, spans; labels + spans | ~14k+ | ▲ ● ◆ | Strong clinical inference, restricted access |
| COVID-QA [41] | Biomedical | CORD-19 papers | Factoid/reasoning; free text | ~2k | ▲ ● ◆ | Small and time-bound |

| Dataset | Domain | Source material | QA format | Size | MH / Sat / Cont | Notes |
|---|---|---|---|---|---|---|
| MedQA (USMLE) [12] | Clinical/medical | Licensing exams | MCQ; single choice | ~12k+ | ▲ (yellow) ● (red) ◆ (red) | Good smoke test, weak discriminative power |
| MedMCQA [13] | Clinical/medical | Indian medical exams | MCQ; single choice | ~194k | ▲ (yellow) ● (red) ◆ (red) | Broad coverage, exam-style leakage risk |
| HealthQA [42] | Consumer health | Web forums | Open-ended; free text | ~1k+ | ▲ (red) ● (yellow) ◆ (yellow) | Realistic register, safety-sensitive |
| LiveQA Medical [43] | Consumer health | NLM consumer queries | Open-ended; free text | ~600+ | ▲ (red) ● (yellow) ◆ (yellow) | Real questions, but small |
| HealthSearchQA [10] | Consumer health | Health search logs | Open-ended; free text | ~3.2k | ▲ (red) ● (yellow) ◆ (yellow) | Real search behavior, harder to memorize |
| MMLU Medical [15] | Medical | Exam questions | MCQ; single choice | ~1k subset | ▲ (yellow) ● (red) ◆ (red) | Heterogeneous and heavily optimized |
| RadQA [27] | Radiology | Radiology reports | Span QA; text spans | ~3k+ | ▲ (red) ● (yellow) ◆ (green) | Domain-specific clinical extraction |
| RealMedQA [44] | Clinical | Clinical guidelines | Open-ended; free text | ~1k | ▲ (yellow) ● (yellow) ◆ (yellow) | Guideline-grounded generation |
| MedHop [37] | Biomedical | Medline + DrugBank | Multi-doc MCQ; single choice | ~2.5k+ | ▲ (green) ● (yellow) ◆ (yellow) | Explicit chaining, but answer cueing |
| MultiMedQA [10] | Clinical/medical | Aggregated multi-source suite | Mixed; various | ~1M+ agg. | ▲ (yellow) ● (yellow) ◆ (red) | Breadth, but inherits legacy leakage |
| HealthBench [29] | Consumer/clinical | Physician-authored scenarios | Open-ended rubric; free text | ~5k | ▲ (yellow) ● (green) ◆ (yellow) | Realistic, rubric-based, unsaturated |
| K-QA [30] | Clinical/medical | Real-world patient questions | Open-ended + NLI metrics; free text | 1,212 | ▲ (red) ● (green) ◆ (yellow) | Completeness/hallucination grading resists saturation |
| MedExQA [45] | Clinical/medical | Five underrepresented specialties | MCQ + explanations; choice + explanation | 965 | ▲ (red) ● (yellow) ◆ (yellow) | Explanation quality beyond accuracy |
| MedREQAL [31] | Biomedical | Systematic reviews | Classification + generation; labels + long answers | 2,786 | ▲ (yellow) ● (yellow) ◆ (yellow) | Evidence-grounded synthesis |
| AfriMed-QA [28] | Clinical/medical | Pan-African med school sources | Open + closed ended; mixed | 15k | ▲ (yellow) ● (yellow) ◆ (yellow) | Broader geography, but partly exam-like |
| LongHealth [32] | Clinical | Long fictional patient cases | MCQ on extraction/negation/sorting; single choice | 400 Qs / 20 cases | ▲ (yellow) ● (green) ◆ (green) | Strong long-context stress test |
| ClinIQLink [33] | Clinical/medical | Expert-verified, source-grounded medical texts | T/F, MCQ, list, short answer, multi-hop; mixed | 4,978 | ▲ (green) ● (green) ◆ (yellow) | Source-grounded and format-diverse |
| BioHopR [38] | Biomedical | PrimeKG | 1-hop/2-hop multi-answer; answer sets | 7,633 | ▲ (green) ● (green) ◆ (yellow) | Explicit KG multi-hop with clear headroom |
| MedQARo [46] | Clinical | Oncology case summaries | Keyword extraction + reasoning; keywords + short text | 105,880 | ▲ (yellow) ● (yellow) ◆ (green) | Large and language-specific |
| MedHopQA | Biomedical | Wikipedia article sets | free text | 1,000 | ▲ (green) ● (green) ◆ (green) | Open-ended chained reasoning with human verification |

*Note: In the compact MH/Sat/Cont column, symbols are ordered as ▲ multi-hop reasoning, ● saturation risk, and ◆ contamination risk. Red = bad, yellow = moderate, and green = good; for saturation and contamination, green means low evaluation risk. Mixed or task-dependent cases are binned to the middle category. Saturation risk estimates how much discriminative headroom a benchmark still offers frontier LLMs. Contamination risk estimates the likelihood that benchmark items, close paraphrases, or answer patterns have entered pretraining, instruction-tuning, or benchmark-optimization corpora. Both are qualitative author assessments for evaluation use, not claims about dataset quality as a training resource.*

1

# Methods

## MedHopQA Benchmark Design Principles

Three design constraints governed every subsequent decision in the construction of MedHopQA. First, each question was required to have at least a two-hop structure: the answer must depend on information distributed across two distinct documents, such that simple retrieval from a single document is insufficient to answer the question reliably. This constraint is what separates a genuine multi-hop benchmark from a single-hop benchmark with longer passages. Second, answers were required to be open-ended free text rather than selections from a closed option set. Multiple-choice formats allow a model to succeed by elimination or surface pattern matching rather than by reasoning; free-text generation forecloses that path. Third, gold annotations were extended beyond a single canonical answer to include all lexically and conceptually valid synonyms, drawn from domain ontologies and verified by human reviewers. Without synonym-inclusive annotation, evaluation penalizes correct answers that happen to use different but equivalent terminology, systematically underestimating model performance on open-ended biomedical tasks. Together, these three constraints define a task that is resistant to the failure modes that limit the discriminative value of existing biomedical QA benchmarks.

## Dataset Creation

The MedHopQA dataset was constructed through a rigorous multi-stage pipeline designed to balance the scalability of AI-assisted generation with the accuracy and validity that biomedical question-answering demands. While artificial intelligence played a key role, for example, in expanding the volume of QA pairs, human expertise remained central at every critical checkpoint in the process. The MedHopQA QA pair design framework is depicted in Figure 1, which is generalizable and applicable to other domains. This framework consists of these steps:

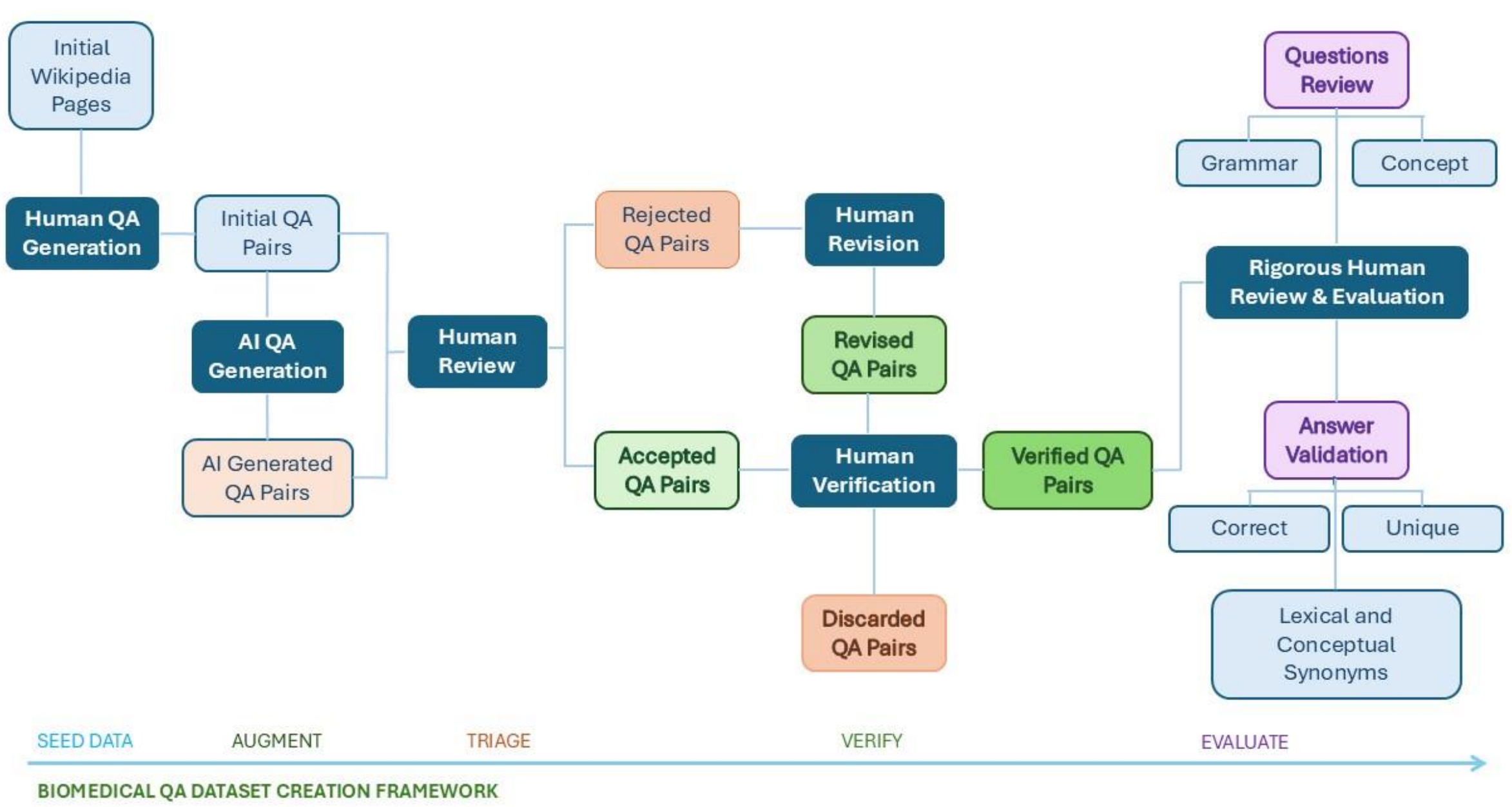

*Figure 1 The dataset creation framework for MedHopQA. Starting with two seed pages from Wikipedia, while AI was used to facilitate steps of this process human review and validation was central to guarantee question and answer validity.*

## 1. Source Material and Seed Page Selection

Wikipedia was selected as the source material by design rather than convenience. Because Wikipedia is universally represented in the pretraining corpora of current LLMs, any question whose answer requires synthesizing information across Wikipedia articles primarily probes reasoning capability rather than memorized recall. A model that has seen both source articles during training still cannot answer a two-hop question correctly without executing the inferential step that connects them; the challenge is relational, not encyclopedic.

The seed dataset was constructed from Wikipedia's curated list of disease pages, which comprises approximately 5,000 entries. Each disease page was treated as a primary document and paired with pages reachable via its outgoing hyperlinks, subject to two filtering rules: outgoing links were capped at 20 per page to avoid combinatorial sprawl, and links appearing in References or External Links sections were excluded, as these tend to point to source citations rather than conceptually related content. Candidate second pages were restricted to those falling within biomedically relevant categories (i.e., diseases, signs and symptoms, genes, chemicals and medications, and related concepts) to ensure that all page pairs supported questions of genuine clinical or biological interest.

The resulting page pairs were shuffled and allocated to sixteen annotators in batches of 20 sets of 20 pairs each. The batch structure was chosen to manage cognitive load while preserving diversity: annotators were free to browse all pairs within an assigned batch and select those they judged most suitable for formulating a valid multi-hop question, rather than being required to produce a question for every pair presented.

## 2. Human Annotation

Sixteen researchers with backgrounds in medicine, biomedicine, bioinformatics, and informatics served as annotators. Each annotator selected page pairs from their assigned batches and formulated a question requiring synthesis of information from both articles. To be accepted, a question had to satisfy five criteria: (1) it addressed medically or biomedically relevant content, including treatment, diagnosis, prognosis, prevention, pathophysiology, or molecular biology; (2) it was factual rather than speculative or probabilistic; (3) it was stable, meaning it was unlikely to become obsolete or require revision as medical knowledge evolves; (4) it was genuinely two-hop, such that answering correctly required content from both source articles and could not be resolved from either alone; and (5) it had a single, unambiguous answer—if the question asked which gene causes a specific disease, exactly one gene could be correct. A representative example is shown in Figure 2.

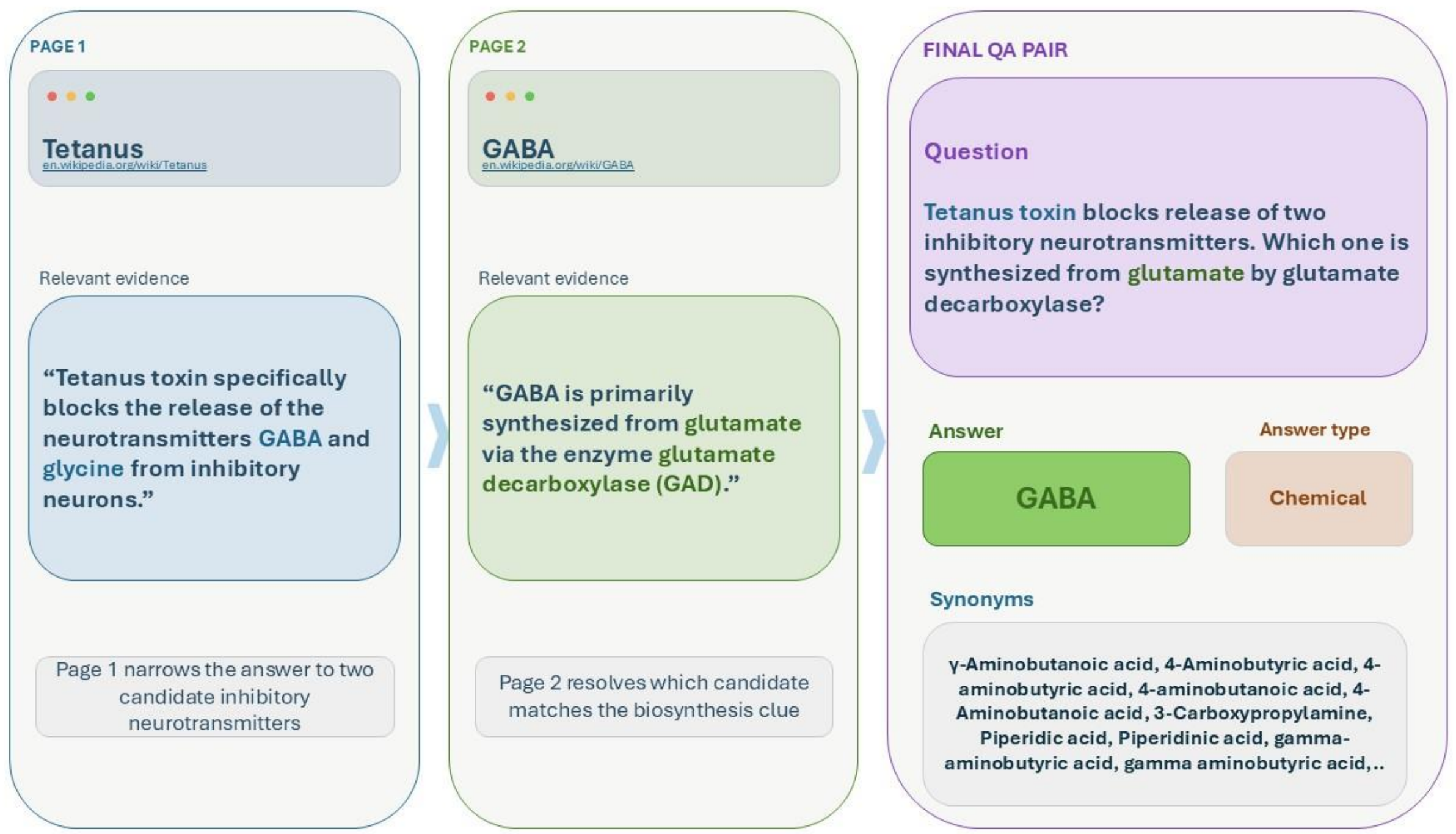

*Figure 2 A multi-hop question-answer pair, typical of the ones found in the MedHopQA dataset.*

Annotators met twice weekly throughout a three-month annotation period. The structure of these sessions evolved as the project progressed. During the first month, when seed question pairs were being generated, discussions focused on developing a shared understanding of multi-hop reasoning and on the range of answer types (i.e., yes/no, named entity or concept, and longer free-form answers). Once the triage phase began in the second month, sessions shifted toward reviewing borderline and rejected questions, which proved to be the most effective training mechanism for calibrating annotator judgment.

Two types of discussion emerged repeatedly and shaped the annotation guidelines. Questions whose intermediate or final answers named very common clinical features (i.e., fever, headache, inflammation) were found to be too easily resolved by general biomedical knowledge, bypassing the intended reasoning path entirely. At the opposite extreme, questions built around highly specific or rare concepts were often trivially solvable by targeted search, even without reading the source articles, due to the specificity of the concept. Valid questions therefore required a middle range of specificity: concepts distinctive enough to demand genuine cross-document synthesis, but not so rare that the answer was effectively given by the question itself. Iterative discussion of concrete examples across the three months progressively sharpened annotators' ability to identify and formulate questions that fell within this range.

### 3. AI Data Augmentation

To scale the dataset beyond what human authorship alone could feasibly produce, an o1-based generation module was used to produce additional QA pairs in parallel with human annotation. The module was prompted with every 100 QA pairs produced by human annotators, using the prompt structure shown in Figure 3, and tasked with generating analogous questions for different disease topics drawn from the same seed page pairs. The source page pair varied with each prompt invocation, ensuring diversity across the generated output. AI-generated pairs entered the same triage pool as human-authored ones and were subject to identical review criteria and acceptance thresholds.

Blinding was a deliberate feature of this design. Annotators were not informed of a question's origin—human or AI—at any stage of triage, and a pair did not advance until two independent reviewers had accepted it. This served two purposes beyond quality control. First, it eliminated systematic bias toward either source: questions could not be accepted or rejected on the basis of who or what produced them. Second, it maintained reviewer

```
You are a helpful assistant for generating multi-hop questions that require two given pages to
answer.

For example, if the first page describes that disease X is caused by gene Y, and the second page
describes that gene Y is located on chromosome Z. A good multi-hop question could be:

"Which human chromosome contains the gene that causes disease X?"

To answer this, a system must first see that gene Y causes disease X from the first page and
that gene Y is located on chromosome Z from the second page.

The final question should be self-contained (decontextualized) and should not be answerable
using only one page. The answer must be short (multiple words are permissible, but sentences are
not desired), unique, and indisputable (e.g., a disease name, a disease treatment name, a gene
name, or yes/no).

Ensure your output is a JSON formatted as follows:

    •   question
    •   reasoning
    •   short answer
```

*Figure 3 The prompt used to create AI generated question-answer pairs*

engagement throughout the triage phase; annotators could not adopt a less rigorous standard for questions they assumed were human-authored.

Early AI-generated questions exhibited two characteristic failure modes that, while requiring rejection, proved instructive. Many could be answered from a single source article, violating the two-hop requirement; others were composite questions embedding two discrete sub-questions, each answerable from one of the two pages, rather than a single question requiring genuine synthesis across both. Because annotators encountered these failure patterns through the normal triage process without knowing their origin, the AI-generated examples functioned as implicit calibration exercises, reinforcing the annotation criteria in concrete terms and sharpening reviewers' ability to detect structurally deficient questions regardless of source.

## 4. Triage and Verification

All QA pairs entered a shared triage pool and were distributed to reviewers concurrently with ongoing generation. Distribution followed strict authorship rules: human-generated pairs were assigned to any annotators except their original author, while AI-generated pairs were distributed equally across the full reviewer pool. No annotator was informed of a pair's origin at any point during triage, ensuring that acceptance and rejection decisions reflected question quality alone.

The triage acceptance rate was low early in the project and improved substantially over time. In the initial weeks, approximately 10% of submitted pairs passed triage; by the end of the annotation period, this figure had risen to 50%. This trajectory reflects two concurrent processes: annotators became more skilled at identifying problematic questions and, in parallel, both human annotators and the AI generation module produced progressively better-calibrated questions as the accepted QA pool grew and iterative feedback accumulated. The triage acceptance rate thus serves as an indirect measure of annotation learning, and the final 50% acceptance rate indicates that the process had reached a stable and well-calibrated state by the time the dataset was completed.

QA Pairs satisfying the criteria explained above advanced, while those failing were returned to the original author for revision. Annotators chose to revise approximately 50% of their submitted queries. AI-authored rejected queries were discarded. In technical terms, the revised QA pairs were redistributed to the researchers that had not seen them before, bringing in fresh eyes to these pairs, and ensuring that each QA pair that graduates to the next stage has been revised and/or validated by at least three researchers. Pairs that passed verification were designated Verified QA Pairs, while those that failed this final check were assigned to Discarded QA Pairs and excluded from the dataset entirely. This two-stage structure—triage followed by verification with disjoint reviewer assignments—was designed to prevent any single annotator's judgment from determining a pair's fate, and to ensure that the cumulative human scrutiny applied to each item increased rather than plateaued at the first review.

## 5. LLM-as-a-Judge Evaluation Framework

Verified pairs entered a structured evaluation phase anchored by a rigorous human review and evaluation process. Four researchers worked intensely at this stage and produced the final set of 1000 QA pairs. This stage also involved the use of specialized LLM-as-judge evaluators which helped structure two parallel assessments: a Questions Review, examining grammar and conceptual accuracy, and an Answer Validation process, examining answer correctness and uniqueness. The evaluators were set up in a debate framework, to evaluate each QA pair from one of four perspectives: 1) Is the question grammatically correct? 2) Is the question conceptually correct? 3) Is the answer a valid answer to the given question? And 4) Is the answer unique, or does there exist another correct answer to the given question?

The outputs of the specialized debaters were presented to an LLM system which was instructed to judge and produce a final report. Four human evaluators carefully reviewed the report and interacted with the system. Human validators revised the QA pairs and re-initialized the evaluation until the QA pair passed all four tests (grammar, coherence, correctness, and uniqueness), as shown in Figure 4.

After the questions and answers were validated, each official short answer was augmented with all known lexical and conceptual synonyms to ensure that valid alternative phrasings were accounted for during evaluation. The synonyms were collected from all related domain ontologies and available resources, such as MONDO (diseases), NCBI Gene (genes/proteins), NCBI Taxonomy (species/organisms), and other supplemental synonyms which were found from Wikipedia or Web. It is important to note that all synonyms underwent significant human oversight because the scope of different ontologies and other vocabularies is different and may not reflect the context of the question, therefore not all existing synonyms available in the ontology may fit each answer.

Taken together, this framework reflects a deliberate design philosophy: AI augmentation was used to achieve the scale necessary for a comprehensive biomedical dataset, but human judgment was embedded at every decision point, from initial triage through final evaluation, to preserve the clinical precision and validity that a domain as sensitive as biomedicine requires.

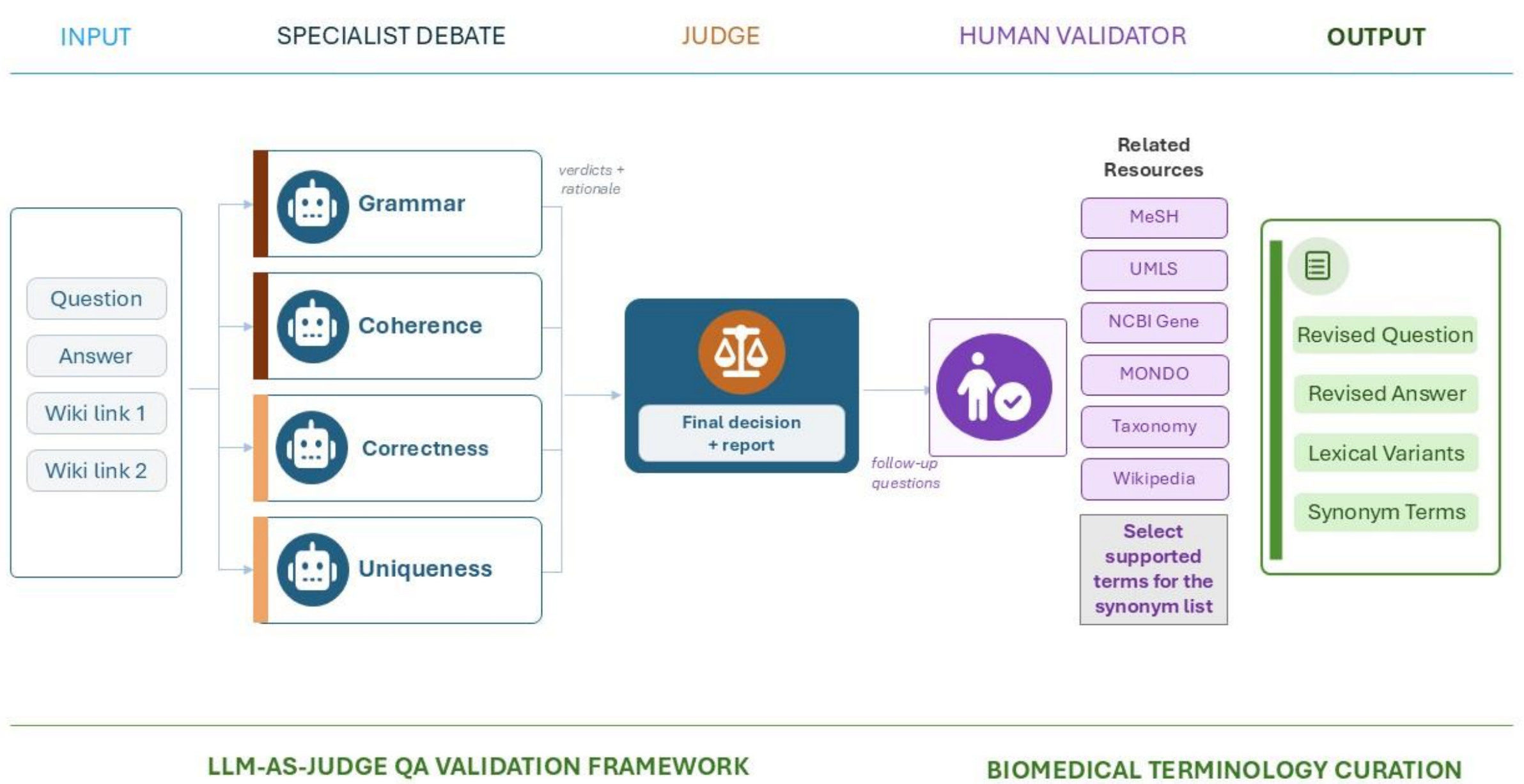

*Figure 4 The LLM-as-a-judge validation framework for QA pairs, and enrichment of the evaluation set with the synonym terms and lexical variants.*

## Dataset evaluation

Evaluating MedHopQA requires accounting for the fact that short biomedical answers may appear in multiple valid surface forms, including abbreviations, lexical variants, and synonymous concept names. At the same time, lexical similarity alone is not sufficient for correct scoring: two answers may look very similar at the string level while referring to distinct biomedical concepts. For example, *type 1 diabetes mellitus* and *type 2 diabetes mellitus* share substantial lexical overlap but are not interchangeable. For this reason, we did not rely on a simple exact-match metric based solely on token overlap.

As described in the Methods section, each ground-truth reference answer is linked to a curated lexicon of acceptable answer forms. For the automated Codabench leaderboard, we implemented a lightweight lexicon-aware string-based metric that we term **Lexical Match**. Under this scheme, a submitted prediction is normalized to reduce variation and compared against both the normalized canonical gold answer and its semantically equivalent variants in the curated lexicon. This approach is more flexible than strict exact matching while remaining transparent, reproducible, and computationally efficient enough for real-time leaderboard scoring. Such flexibility is particularly important in biomedical QA, where correct answers may differ in capitalization, punctuation, abbreviation, lexical form, or use of colloquial versus technical terminology, for example, *high blood pressure* versus *hypertension*.

To preserve blind evaluation and reduce leaderboard gaming, the Codabench setup also uses hidden questions. Specifically, the MedHopQA set of 1,000 evaluation questions is embedded within a larger 10,000-question package that participants can download and run inference on, while only the hidden MedHopQA subset is used for official scoring. This design reduces manual overfitting to visible examples, makes repeated submission-based answer engineering more difficult, and provides a more reliable estimate of model generalization. It also offers partial protection against benchmark contamination by limiting direct exposure of the scored test items.

As a secondary evaluation, we performed concept-level validation using MedCPT [47], a specialized biomedical representation model trained on large-scale PubMed search logs. MedCPT generates biomedical text embeddings that can be used to measure the semantic similarity between a pair of input texts. This analysis is designed to determine whether the predicted answer and the reference answer refer to the same underlying biomedical concept, even when their lexical forms differ beyond what can be captured by string normalization or curated synonym lists alone. Because this procedure requires embedding generation and large-scale similarity search over domain lexicons, it is computationally more demanding and is therefore performed locally rather than within the Codabench evaluation pipeline.

More concretely, this evaluation can be viewed as a two-stage normalization and verification process: (i) mapping free-text predictions into a structured biomedical concept space, and (ii) verifying equivalence against curated answer sets within that space. To ensure appropriate handling of heterogeneous answer types, concept-level validation is divided into three sub-evaluations:

1. Non-semantic answer types (yes/no, numerical, chromosome location). These categories require exact or near-exact formatting rather than semantic interpretation. For these cases, we apply the same evaluation procedure as the lexical match metric. Concept-level modeling is not applied because semantic similarity is either ill-defined (e.g., numeric values) or inappropriate (e.g., chromosome band notation).
2. Entity-based answer types (disease, gene/protein, chemical, anatomical entity, sign/symptom, species). These categories correspond to entities that are well represented in biomedical ontologies and databases. For these types, we perform lexicon-constrained semantic normalization, which maps predictions to canonical concepts before evaluation.
3. Open or weakly structured answer types (other, description, procedure, person, date, cell type, treatment). These categories lack a single authoritative ontology or exhibit high variability in expression. For these cases, we use direct embedding-based similarity without lexicon projection.

For entity-based answers, we leverage curated lexicons derived from domain resources (e.g., MONDO for diseases, NCBI Gene for genes/proteins, MeSH or related resources for chemicals and anatomy). The key idea is to project both predictions and gold answers into a shared concept space defined by the lexicon, thereby reducing sensitivity to surface variation while maintaining strict semantic discrimination.

The procedure is as follows:

1. Embedding of prediction. The model prediction is converted into a dense vector representation using MedCPT.
2. Lexicon retrieval. The embedding is used to perform nearest-neighbor search over the lexicon corresponding to the answer type. Each lexicon entry (typically a canonical name or synonym) is also represented as a MedCPT embedding.

3. Nearest-neighbor projection. The lexicon entry with the highest cosine similarity to the prediction embedding is selected. This step effectively maps the free-text prediction to its closest canonical concept within the constrained vocabulary.
4. Normalization of projected prediction. The selected lexicon string is normalized using the same preprocessing pipeline applied in lexical matching (e.g., lowercasing, punctuation handling).
5. Exact match against curated synonyms. The normalized projected prediction is compared against the normalized set of manually curated synonyms for the reference answer. The prediction is marked correct if and only if an exact match is found.

This procedure serves two complementary purposes:

- Smoothing over lexical variation: Different surface forms (e.g., abbreviations, alternate spellings) are collapsed via embedding similarity and lexicon projection.
- Preserving semantic precision: Because the mapping is constrained to a lexicon, the prediction must be closer to the correct concept than to any competing concept in the ontology, reducing false positives due to broad semantic similarity.

For answer types that are not well covered by a single ontology (e.g., descriptions or procedures), we instead perform pairwise semantic comparison between the prediction and all curated reference variants:

- The prediction and each reference synonym are converted to MedCPT embeddings.
- Cosine similarity is computed between the prediction embedding and each reference embedding.
- The prediction is marked correct if and only if the maximum similarity exceeds a predefined threshold (0.7).

This formulation treats correctness as a semantic proximity criterion rather than exact equivalence. The threshold was selected empirically to approximately balance tolerance to paraphrasing against the risk of accepting semantically related but incorrect answers.

## Results and Discussion

### Dataset composition

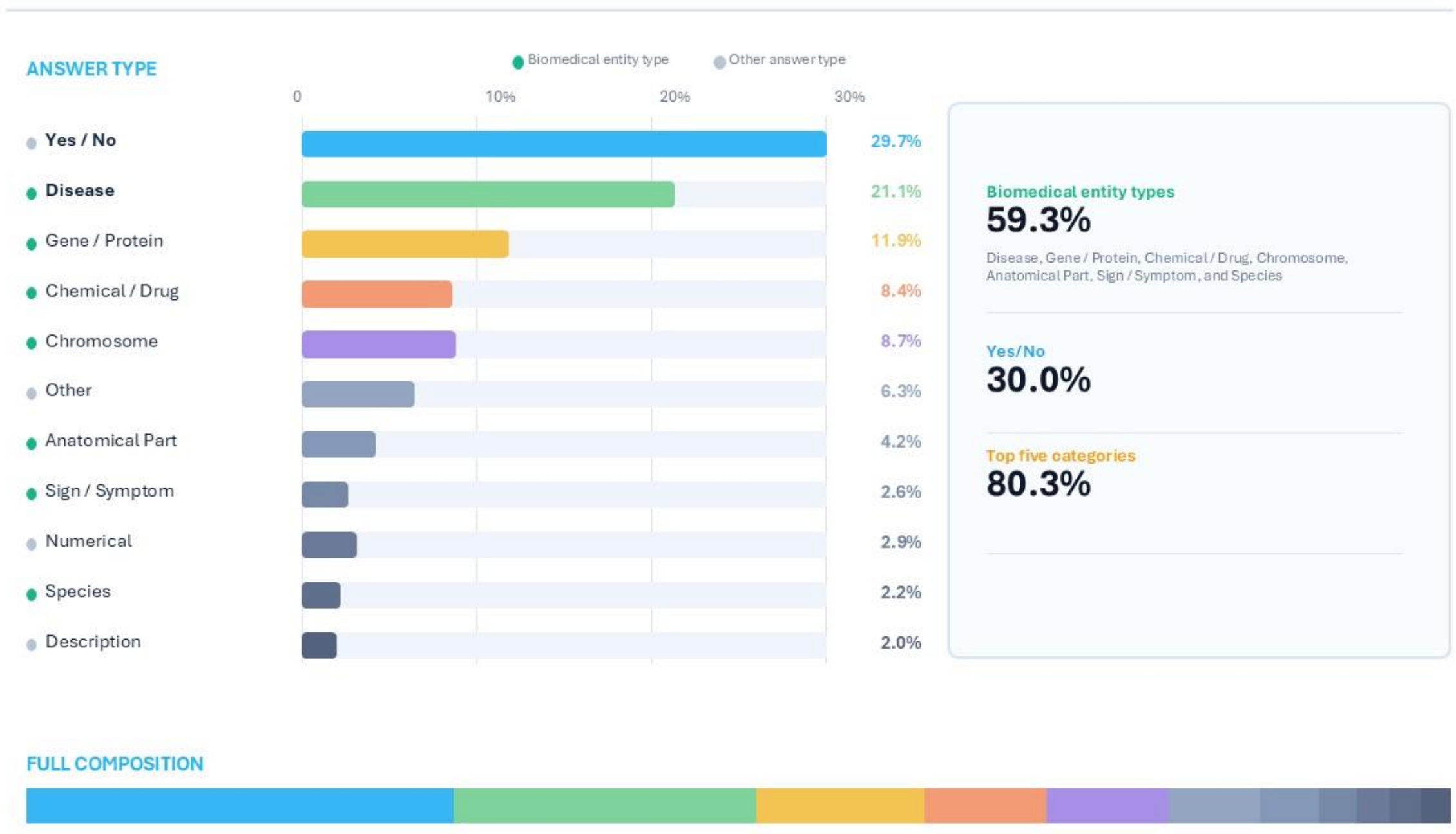

*Figure 5 Distribution of answer types in the MedHopQA dataset*

The distribution of answer types (Figure 5), provides a concise characterization of the dataset's scope and difficulty. Yes/no questions constitute the largest single category (30%). Biomedical entity-based answers collectively account for 59.3% of the dataset, with Disease as the most frequent category (21.3%), followed by Gene/Protein (11.8%), Chemical/Drug and Chromosome (8.4% and 8.7%). This distribution reflects our deliberate emphasis on disease-centered reasoning grounded in core biomedical entities.

Answer length is tightly constrained: the average answer is approximately two words, with most consisting of one- or two-word responses. Single-word answers are dominated by yes/no responses and core biomedical entities (gene, disease and chemical names). Two-word answers also cover the disease and chemical categories, but also cover chromosomes and anatomical terms. Two-word answers often follow compositional patterns, such as *adjective/specifier* + *noun* (e.g., *Alzheimer's disease*).

The dataset is constructed from 1300 unique Wikipedia pages. The collection is strongly disease-centered, covering a wide range of conditions, including rare/congenital disorders, developmental anomalies, cancers, infections, and clinical findings. Significant secondary topics include genes/proteins (e.g., receptors and enzymes), drugs/chemicals, anatomical entities, and clinically-relevant pathogens (viral, bacterial, and protozoan).

The disease distribution (Figure 6) further emphasizes rare and complex conditions, particularly genetic syndromes, metabolic disorders, and hereditary neurological diseases. Notably, nearly three-quarters of the diseases are assigned to multiple categories (e.g., neurological + metabolic + genetic, or cancer + hematological + dermatological). The high rate of multi-category diseases reflects the multi-system nature of many rare diseases. This multi-category structure contributes directly to the need for multi-hop reasoning, as answering questions often requires integrating information across systems rather than within a single domain.

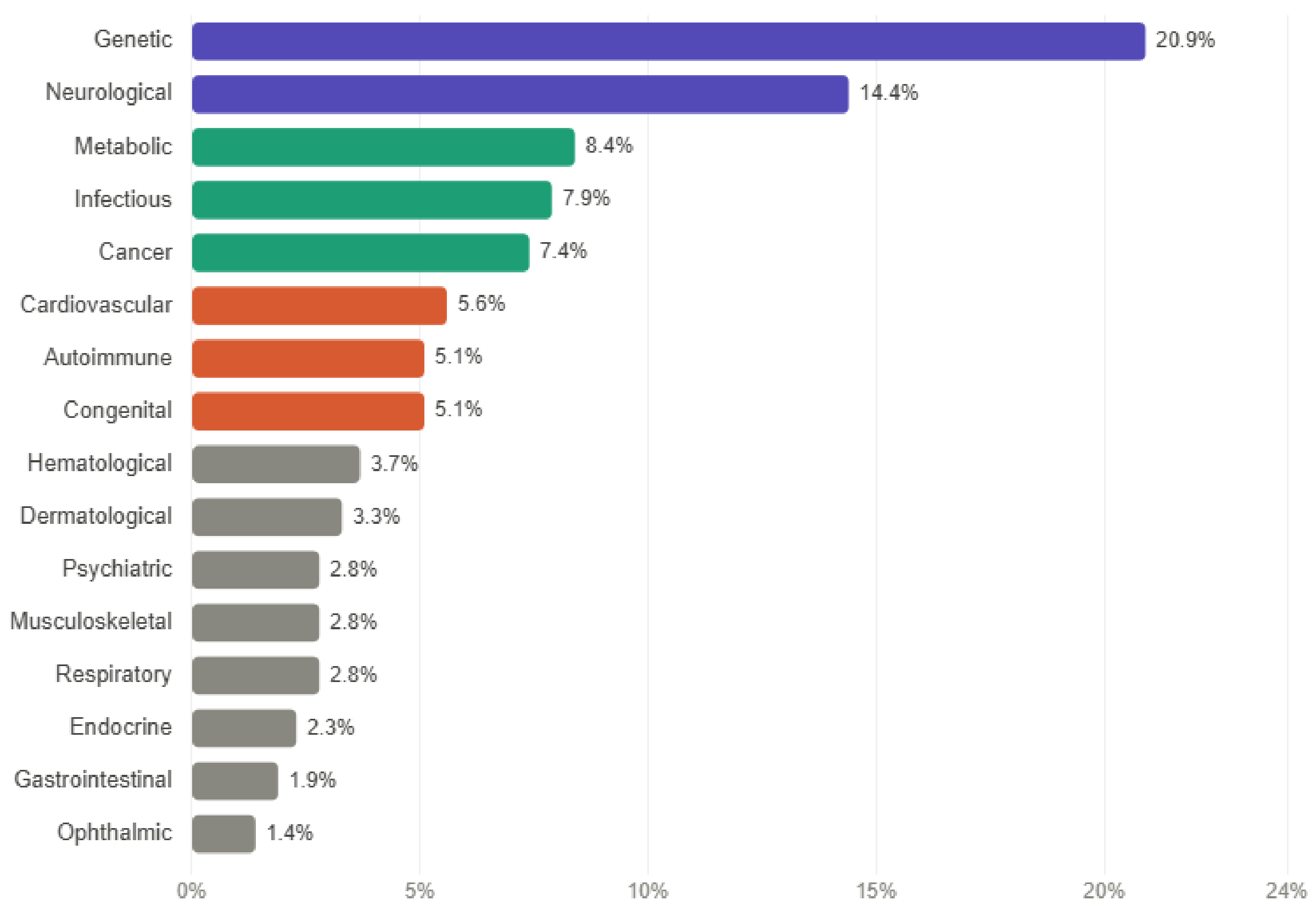

*Figure 6 Disease category distribution in the MedHopQA dataset. ▲ high-tier, ▲ mid-tier, ▲ moderate, ▲ small.*

## Evaluation

We evaluated four frontier LLM models under a controlled zero-shot setting: GPT-4o, GPT-5.1, Claude Sonnet 4.5, and Gemini 2.5 Pro. This evaluation is not intended as benchmarking, but rather as a controlled probe of dataset properties. Each model received only the question and a short instructive prompt. No supporting documents or source URLs were provided. This design isolated the intrinsic ability of models to perform multi-hop reasoning without external retrieval; performance differences can therefore be interpreted as reflecting model reasoning and internal knowledge representations rather than retrieval or pipeline engineering.

This evaluation setup mirrors the constraints used in the MedHopQA shared task at BioCreative IX. Here the workshop participants were likewise not provided with the Wikipedia source documents, though participants were free to prepare and retrieve their own set of supporting documents.

| | | Lexical Match Evaluation | | | | Concept-level Evaluation | | | |
|---|---|---|---|---|---|---|---|---|---|
| **Answer type** | **%** | **GPT-4o** | **Claude Sonnet 4.5** | **GPT-5.1** | **Gemini 2.5Pro** | **GPT-4o** | **Claude Sonnet 4.5** | **GPT 5.1** | **Gemini 2.5Pro** |
| Yes/No | 29.7% | 69.4% | 83.5% | 86.5% | 80.8% | 69.4% | 83.5% | 86.5% | 80.8% |
| Disease | 21.1% | 53.1% | 71.6% | 73.0% | 77.3% | 63.0% | 75.4% | 81.5% | 80.1% |
| Gene/Protein | 11.9% | 55.5% | 71.4% | 61.3% | 70.6% | 60.5% | 72.3% | 80.7% | 77.3% |
| Chemical/Drug | 8.4% | 81.0% | 82.1% | 82.1% | 85.7% | 84.5% | 84.5% | 89.3% | 89.3% |
| Chromosome | 8.7% | 58.6% | 66.7% * | 83.9% | 77.0% | 58.6% | 66.7% * | 83.9% | 77.0% |
| Anatomical Part | 4.2% | 57.1% | 71.4% | 66.7% | 73.8% | 71.4% | 76.2% | 83.3% | 83.3% |
| Sign/Symptom | 2.6% | 57.7% | 73.1% | 69.2% | 73.1% | 57.7% | 76.9% | 73.1% | 73.1% |
| Numerical | 2.9% | 20.7% | 34.5% | 48.3% | 31.0% | 20.7% | 34.5% | 48.3% | 31.0% |
| Species | 2.2% | 36.4% | 86.4% | 86.4% | 81.8% | 50.0% | 90.9% | 90.9% | 86.4% |
| Description | 2.0% | 40.0% | 35.0% | 65.0% | 65.0% | 80.0% | 70.0% | 80.0% | 80.0% |
| Other | 6.3% | 42.9% | 57.1% | 66.7% | 66.7% | 82.5% | 84.1% | 90.5% | 88.9% |
| **Overall** | **100.0%** | 59.1% | 73.2% * | 76.0% | 75.8% | 66.3% | 77.1% * | 83.4% | 79.7% |

*Other = other + procedure + person + date + cell type + treatment*
**adjusted to account for formatting issue*

0 | +5 | +10 | +20 | +40

*Figure 7 Accuracy of zero shot evaluation for leading LLM models*

Performance varies substantially across answer types, highlighting the diagnostic value of the dataset (Figure 7). GPT-5.1 achieves the highest overall accuracy (83.4%), followed by Gemini 2.5 Pro (79.7%), Claude Sonnet 4.5 (77.1%), and GPT-4o (66.3%). This ranking is preserved under macro-averaging across answer categories, indicating that results are not driven solely by the dominant yes/no subset.

Several categories are handled relatively consistently across models. *Chemical* questions show the highest and most stable performance across both evaluation settings (84.5% and 89.3%), suggesting that these items exhibit lower variability and are more discriminable than other types. *Anatomical* questions also exhibit comparatively strong performance (71.4% to 83.3%). Smaller categories, grouped under *Other* (e.g., *procedure, date, cell type,* and *treatment*), show high scores, but reliability is limited due to small sample sizes.

In contrast, performance on *disease* (63.0% to 81.5%) and *gene/protein* questions (60.5% to 80.7%) is more variable. These categories require fine-grained discrimination between closely related biomedical concepts with substantial lexical overlap, making them more sensitive to both reasoning and normalization errors. A similar pattern is observed for *sign/symptom* questions (57.7% to 76.9%), where clinically similar concepts introduce ambiguity.

*Numerical* questions represent the most challenging category overall. *Chromosome* location questions also show sensitivity to formatting; correcting these formatting issues substantially improved Claude Sonnet's performance in this category.

These patterns highlight two distinct sources of difficulty captured by the dataset:

- **Semantic discrimination** (e.g., *disease, gene/protein*), where models must distinguish closely related concepts
- **Normalization precision** (e.g., *numerical, chromosome*), where exact formatting is required

Importantly, the evaluation framework interacts differently with these categories. Entity-based answers are more sensitive to synonym coverage and lexical variation, whereas numerical and structured answers depend primarily on exact matching. Although the curated synonym lexicon reduces variability, it is not exhaustive, and some residual mismatches likely reflect incomplete coverage rather than true reasoning errors.

At the question level, 534 of 1,000 questions are answered correctly by all four models, while 77 are missed by all. This distribution suggests a shared "easy core" alongside a persistent hard subset.

Concept-level validation complements the lexical match metric by operating at a higher level of abstraction. While lexical matching emphasizes precision under controlled normalization, MedCPT-based evaluation captures semantic equivalence under variation in expression. As observed in the results, concept-level scores are typically higher, reflecting successful recovery of correct answers that differ in surface form. At the same time, the lexicon-constrained design ensures that this increased sensitivity does not come at the cost of collapsing distinct biomedical concepts into a single equivalence class.

Together, these properties make the MedCPT evaluation particularly suitable for biomedical QA, where correctness depends on identifying the right concept, not merely producing a string with high lexical overlap.

## Error Analysis

We manually reviewed 200 MedCPT decisions and verified an overall agreement of 75.5% (151/200), however the nature of the error is different as shown in Figure 8. The dominant error is false rejection of correct answers, not false acceptance of wrong ones: 28 of the 49 errors (57.1%) are cases where the answer was actually correct, but MedCPT did not validate it, whereas 19 errors (38.8%) are cases where MedCPT accepted an incorrect answer. False negatives were mostly related to an unrecognized answer, such as incomplete concept name (partial string, which fits in with the context), more elaborate answer (concept-name is part of the given answer), or incomplete synonym-list, and they appear to reflect normalization limitations. Performance is highest for *gene/protein* (96.7%), *anatomical* (95.8%), and *sign or symptom* (93.8%) answers, and remains solid for *chemical* (82.8%), suggesting that MedCPT is most reliable when the target answer maps cleanly to a canonical biomedical entity. The error profile is also category-specific: *disease* and *chemical* errors are mostly missed correct answers, while the category: *other* contributes most of the false positives, indicating that MedCPT is much more reliable for canonical biomedical entities. In Figure 8 we also identify categories such as *person* or *date* which were previously grouped with *other*. This pattern suggests that the main opportunity for improvement is twofold: first, expanding normalization and synonym coverage for *diseases*, and *species*, and second, tightening concept-level matching to reduce false positives in categories such as *other*, *person*. These results support using MedCPT as a useful concept-level validator.

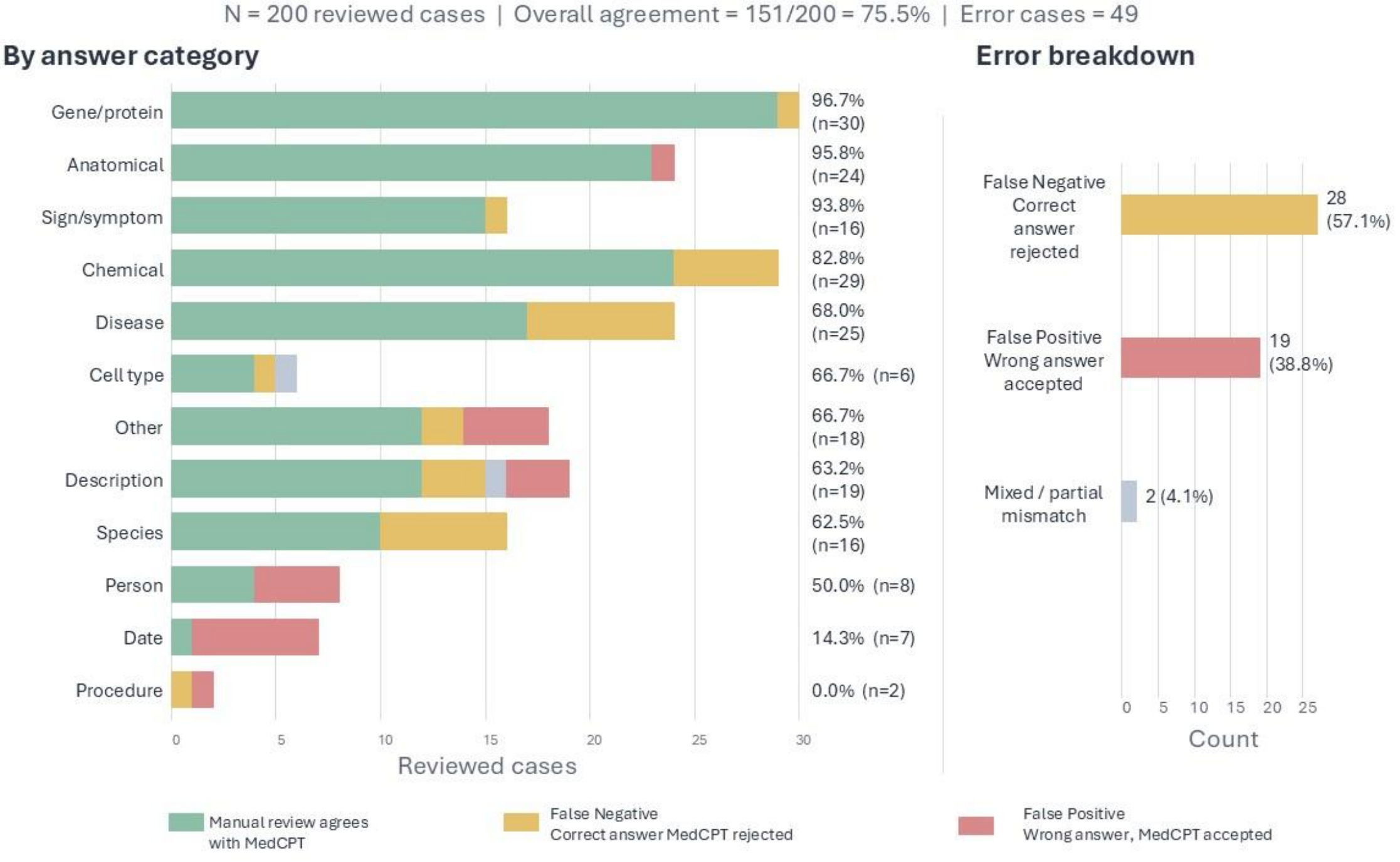

*Figure 8 Error analysis of 200 answers, and their respective evaluations with MedCPT*

## Discussion & Limitations

Reliable evaluation of biomedical LLMs requires benchmarks that both differentiate among high-performing systems and reflect the types of reasoning these systems are expected to support in practice. Many existing biomedical QA benchmarks prioritize formats that enable straightforward comparison (e.g., multiple-choice or span extraction), but these formats can reduce discriminative headroom and allow performance to be driven by localized evidence or memorization effects. In contrast, open-ended, multi-hop benchmarks remain relatively limited in the biomedical domain.

MedHopQA addresses this need with a dataset of 1,000 disease-centered, multi-hop question–answer pairs constructed through a multi-stage human–AI workflow and evaluated within the BioCreative IX shared task framework. By requiring integration of information across two distinct source documents and supporting open-ended answers evaluated with synonym sets, MedHopQA creates conditions under which simple pattern matching is insufficient and multi-step reasoning is more directly engaged. The evaluation results indicate that, even across frontier models, performance varies substantially across question types and remains sensitive to both conceptual distinctions and output representation constraints. LLMs demonstrate that multi-hop biomedical reasoning remains a genuine challenge.

Beyond its immediate use as a benchmark, MedHopQA provides a framework for constructing evaluation datasets that balance scalability with domain fidelity. The combination of targeted seed curation, AI-assisted question generation, and iterative human validation offers a practical, reusable template for benchmark development that requires compositional reasoning while maintaining biomedical accuracy. The framework's central design principle, that AI augmentation should serve human annotation rather than replace it, proved essential to maintaining the factual precision required in the sensitive biomedical domain. This workflow may be applicable to other biomedical subdomains, although adaptation will require domain-specific expertise and curation strategies.

This work has several limitations that should be considered when interpreting the results. First, the dataset is derived exclusively from English-language Wikipedia, and therefore reflects its coverage and potential biases. Periodic updates to the source materials will be necessary to maintain benchmark relevance.

Second, the dataset is intentionally disease-centered. While this focus targets a high-priority application, it limits coverage of other important domains, such as pharmacology, genomics, clinical procedures, and public health. Extension to these areas represents natural but non-trivial future work, as each area will require domain-specific curation strategies and expert annotation.

Third, the current evaluation framework assesses answer correctness but not reasoning quality. The combination of lexical matching and concept-level validation provides robust outcome-based evaluation, but does not capture intermediate reasoning steps. Future extensions could incorporate structured reasoning annotation or expert evaluation of reasoning tracts to better assess multi-hop inference.

Finally, all annotators were affiliated with a single institution and shared overlapping disciplinary backgrounds in biomedicine and computational science. While this ensured domain expertise and consistency, it may have introduced systematic biases in question formulation or difficulty calibration. An annotator pool with a wider distribution of geographic locations and disciplines could improve coverage and robustness.

## Conclusions

As generative AI models are increasingly used in biomedical research and clinical decision support, evaluation frameworks must capture not only whether answers are correct, but also the conditions under which systems succeed or fail. MedHopQA is intended as a step toward this goal. It provides a benchmark that exposes multiple dimensions of difficulty within biomedical question answering, and it offers a practical framework for developing future benchmarks with saturation and contamination resistance as design constraints.

The MedHopQA dataset is available at: CodaBench Link.

## Acknowledgments

This research was supported by the Intramural Research Program of the National Institutes of Health (NIH). The contributions of the NIH authors are considered Works of the United States Government. The findings and conclusions presented in this paper are those of the authors and do not necessarily reflect the views of the NIH or the U.S. Department of Health and Human Services.